# Consensual Collaborative Training and Knowledge Distillation based Facial Expression Recognition under Noisy Annotations

Darshan Gera [1], S. Balasubramanian [2]

[1]Assistant Professor, DMACS, SSSIHL, Brindavan, Bengaluru, Karnataka, India

[2]Associate Professor, DMACS, SSSIHL, Puttaparthi, Anantpur, A.P., India

[1]darshangera@sssihl.edu.in, [2]sbalasubramanian@sssihl.edu.in

*Abstract -* Presence of noise in the labels of large scale facial expression datasets has been a key challenge towards Facial Expression Recognition (FER) in the wild. During early learning stage, deep networks fit on clean data. Then, eventually, they start overfitting on noisy labels due to their memorization ability, which limits FER performance. This work proposes an effective training strategy in the presence of noisy labels, called as Consensual Collaborative Training (CCT) framework. CCT co-trains three networks jointly using a convex combination of supervision loss and consistency loss, without making any assumption about the noise distribution. A dynamic transition mechanism is used to move from supervision loss in early learning to consistency loss for consensus of predictions among networks in the later stage. Inference is done using a single network based on a simple knowledge distillation scheme. Effectiveness of the proposed framework is demonstrated on synthetic as well as real noisy FER datasets. In addition, a large test subset of around 5K images is annotated from the FEC dataset using crowd wisdom of 16 different annotators and reliable labels are inferred. CCT is also validated on it. State-of-the-art performance is reported on the benchmark FER datasets RAFDB (90.84%), FERPlus (89.99%) and AffectNet (66%).

**Keywords —** Collaborative training, Crowd-sourcing, Knowledge distillation, Facial expression recognition, Noisy annotation.

## I. INTRODUCTION

Recognizing expressions in faces plays a vital role in communication and social interaction, analysing mental related illness like depression, measuring attentiveness in student-teacher interaction etc. Due to the resurgence of deep neural networks (DNNs) and the availability of large datasets, automatic facial expression recognition (FER) systems have received a lot of attention recently [1]. Such systems have plethora of applications including human-computer interaction [2], intelligent tutoring [3], automatic driver alert [4], mental health analysis [5] and computer animation [6].

Traditional works like [7]-[10] focused on training machines for FER through examples collected in a controlled (in-lab) environment. Examples of such in-lab datasets are CK+ [11]-[12], Oulu-CASIA [13] and JAFFE [14]. These datasets are small in size, and annotations are readily available. However, machines trained on small in-lab datasets do not generalize well to real-world scenarios. Consequently, large datasets collected from real-world scenarios (called as in-the-wild datasets) like AffectNet [15], FERPlus [16] and RAFDB [17]-[18] have been made available. However, these datasets contain noisy annotations that are primarily caused by (i) crowd sourcing, (ii) ambiguity in expressions, (iii) poor quality of images due to variations in illumination, pose, occlusion, and low resolution and (iv) automatic annotations obtained by querying web using keywords [19]. Such noisy annotations can significantly affect the performance of DNNs [20]-[22]. Therefore, it is important to handle noisy annotations while training.

Two approaches have been adopted by bulk of the methods to deal with noisy annotations. One is to correct the noisy labels by estimating the noise transition matrix [23]-[24]. The other is to identify the noisy labels and suppress their influence during training [25]-[26]. The former approach is generally underperforming since it is difficult to accurately estimate the noise transition matrix. The later identifies clean labels as ones with low loss during early part of the training, relying on the fact that DNNs fit clean labels first before overfitting on the noisy labels [20]. Since hard samples do not have low loss during early part of the training, they will get clubbed with noisy samples. This is detrimental to generalizability of the approach. Further, the later approach requires prior information about the noise distribution in the data, which is not available in the real-world scenario.

In this work, the problem of noisy annotations is approached differently. Three networks are co-trained jointly. Each network is facilitated to learn expressions in faces through a supervision loss. The influence of noisy labels on each network is suppressed by forcing each network to be consistent with the other. This is achieved through a consistency loss. A dynamic transition scheme is used to gradually move from supervision loss to consistency loss during training. This facilitates influence of clean labels during early part of the training and prevent overfitting to noisy labels during later part of the training. CCT does not require noise distribution to be known in advance, and hence fits very well in real-world scenarios. CCT is presented in detail in section III. A simple knowledge distillation strategy is proposed for inference wherein a single network is distilled from the co-trained three networks and used for inference. In addition, as a contribution towards FER research community, a large test subset of around 5K images is annotated and released from the test set of FEC dataset [27] which is made available by google recently. The details are presented in section IV. In summary, our contributions are as follows:

1. CCT framework for FER with noisy annotations without any assumption about noise distribution.
2. A knowledge distillation scheme to distill a single network for inference from the co-trained three networks.
3. Annotation of a large test subset of around 5K images from the FEC test dataset.
4. Robustness of the proposed framework in the presence of synthetic label noise on RAFDB [17]- [18], FERPlus [16] and AffectNet [15] datasets.
5. Robustness of the proposed method on real noisy datasets including AffectNet and our curated FEC.
6. Newest state-of-the-art (SOTA) performance on standard FER datasets including RAFDB, FERPlus and AffectNet demonstrating CCT's utility as a general purpose robust FER learning framework.

## II. RELATED WORK

Handling noisy annotations in DNNs is challenging since DNNs tend to over fit them easily [20]. Many strategies have been proposed viz. small loss trick [25]-[26], [28], robust losses [29]-[30], label cleansing [31], weighting [32], meta-learning [33], ensemble learning [34], and others [35]-[36]. Earlier approaches [23]-[24] correct labels by estimating the noise transition matrix. For example, [37] considers labels as probability distributions over the classes and iteratively corrects these distributions. However, obtaining an accurate noise transition matrix is not possible in real-world scenario. The current trend is to rely on low loss samples with or without co-training [25]-[26], [38]-[40] due to the fact that low loss samples are associated with clean labels [26]. These methods are also conservative in their later part of training since DNNs tend to memorize noisy labels [20]-[21] eventually.

In [26], a teacher net guides the student net to choose samples with low loss, assuming it has access to a clean dataset, and subsequently learning from the dataset. A clean dataset may not be always available. In the absence of clean dataset, the method depends on a pre-defined curriculum, which is generally difficult to conceive for a DNN. The method suffers from self-accumulated error. Decoupling [38] attempts to overcome this by training two networks, each of which will independently update based on their low loss samples on whom the network predictions disagree with the other network's predictions. Such low loss samples are likely to be associated with clean labels. However, since noise can span across the entire label set, the disagreement region predominantly contains noisy labels, and hence in such cases, the method becomes less effective. Coteaching [25] and Co-teaching+ [28] tackle this by updating each network based on small loss samples selected by the other network, thereby incorporating peer learning. The two ideas differ in the sense that Co-teaching+ adopts low loss sample selection based on disagreement to keep both the networks sufficiently diverged. Main limitation of the aforementioned methods is that they depend on accurate estimation of noise distribution in the data.

Contrary to the disagreement principle followed in [38, 28], JoCoR [39] selects low loss samples based on agreement between two networks, enforced by a joint loss. However, JoCoR requires noise rate to be known in advance. NCT [40] performs mutual training but does not rely on low loss samples explicitly. During training, along with self supervision, it forces one network to imitate the other. Also, to enforce complete training, it flips target labels randomly based on an adaptive target variability rate that it computes. Further, it does not require noise rate to be known in advance. Though a lot of methods are discussed above for handling noisy labels, a large number of them have been tested only on datasets like MNIST, CIFAR10 etc. Literature on handling noisy labels in FER is scarce. In [41], each sample is assigned multiple inconsistent human and machine predicted labels, and subsequently a network is trained to discover the true label. In [19], self-cure network (SCN) learns the importance of each sample using a self-attention module for loss re-weighting. Low weight samples are treated as noisy and also relabeled if predicted probability is above a certain threshold. Both these methods focus on correcting the noisy labels, which comes with uncertainty.

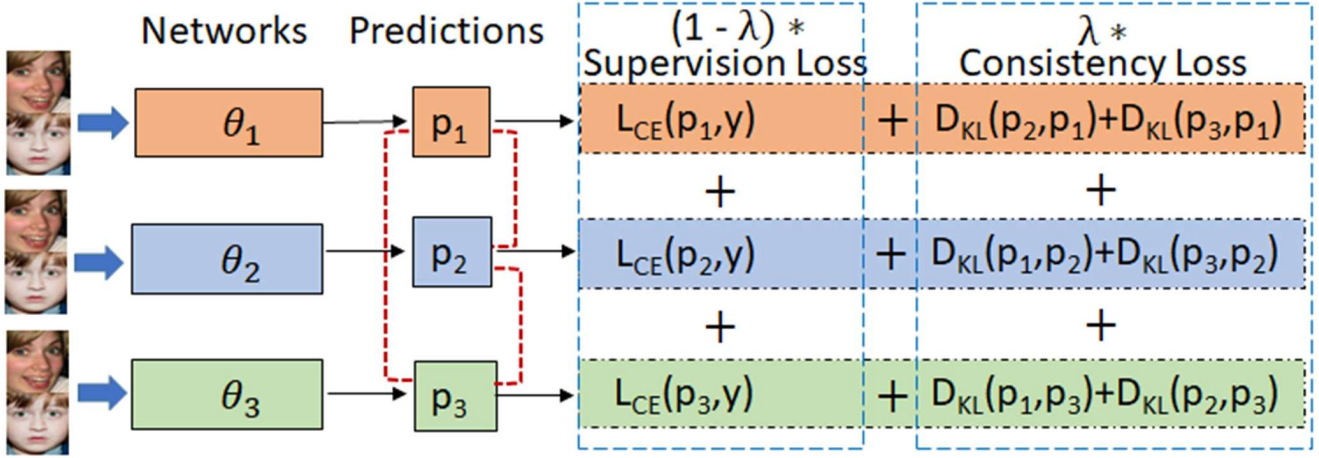

Fig. 1 CCT involves training three networks θ₁; θ₂ and θ₃ jointly using a convex combination of supervision loss and consistency loss. Consensus is built by aligning the posterior distributions (shown as dotted red curves between p₁, p₂ and p₃) using consistency loss. Dynamic weighting factor (λ) that balances both the losses is described in Fig. 2 and Eq. 4

### III. PROPOSED CCT

*A. Overview*

The proposed CCT follows the principle of consensus based collaborative training called Co-Training [42]. It uses three networks to learn robust facial expression features in the presence of noisy annotations. Inspired by [39]-[40], [43], three networks with identical architecture, but different initializations, are trained jointly using a convex combination of supervision loss and consistency loss. Different initializations promote different learning paths for the networks, though they have same architecture. This subsequently reduces the overall error by averaging out individual errors due to the diversity of predictions and errors. In the initial phase of training, networks are trained using supervision loss. This ensures that clean data is effectively utilized during training since DNNs fit clean labels initially [20]-[21]. Further, to avoid eventual memorization of noisy labels by individual DNNs, gradually, as the training progresses, more focus is laid on consensus building using consistency loss between predictions of different networks. Building consensus between networks of different capabilities ensures that no network by itself can decide to overfit on noisy labels. Further, it also promotes hard instance learning during training because the networks are not restricted to update based on only low loss samples, and further that, as the training progresses, they must agree. The trade-off between supervision and consistency loss is dynamically balanced using a Gaussian like ramp-up function [40]. Further, the proposed CCT does not require noise distribution information, and it is also architecture independent. Fig. 1 delineates the proposed CCT framework and Algorithm 1 enumerates the pseudo-code for CCT training. Note that CCT is different from JoCoR [39] in that CCT does not need to know noise rate in advance, unlike JoCoR. In addition, unlike JoCoR, CCT dynamically balances between different losses. CCT also differs from NCT [40] wherein target variability is introduced to avoid memorization in later stages. CCT implicitly avoids memorization because it has three networks collaborating with each other through the consistency loss. It does not require target variability function. Also, supervision is independent in NCT while it is done jointly in CCT. Overall, CCT demonstrates superior performance on both synthetic and real noisy labels, as reported in section IV.

*B. Problem Formulation*

Given C expression categories, let D=$\{(x_i, y_i)\}_1^N$ be the dataset of N training samples where $x_i$ is the ith input facial image with expression label $y_i \in \{1,2,..,C\}$. CCT is formulated as a consensual collaborative learning using three networks parametrized by $\{\theta_j\}_{j=1}^3$. The learning is achieved by minimizing the loss L given by:

$$L = (1 - \lambda) * L_{sup} + \lambda * L_{cons} \quad (1)$$

where $L_{sup}$, $L_{cons}$ and λ are described below.

*1) Supervision Loss*

Cross-entropy (CE) loss is used as supervision loss to minimize the error between predictions and labels. Let $p_j$ denote the prediction probability distribution of $j^{th}$ network. Then,

$$L_{sup} = \sum_{j=1}^{3} L_{CE}^j(p_j, y) \quad (2)$$

where y is the groundtruth vector.

*2) Consistency Loss*

We use Kullback-Leibler divergence (DKL) to bring consensus among predictions of different networks by aligning their probability distributions.

$$L_{cons} = \sum_{j=1}^{3} \sum_{\substack{k=1 \\ k \neq j}}^{3} D_{KL}^{(j)}(p_k \parallel p_j) \quad (3)$$

*3) Dynamic Balancing*

The dynamic trade-off factor between $L_{sup}$ and $L_{cons}$ is computed as in [40]. Specifically,

$$\lambda = \lambda_{max} * e^{-\beta*(1-\frac{e}{e_r})^2} \quad (4)$$

where β determines the shape of the Gaussian like ramp-up function, $\lambda_{max}$ refers to maximum value of λ, e is the current epoch and $e_r$ is epoch (ramp-up length) at which λ attains its maximum value ($\lambda_{max}$). When λ is small, supervision loss dominates. As training progresses, e approaches $e_r$ which pushes λ towards $\lambda_{max}$, allowing consistency loss to take over. Fig. 2 displays the dynamic balancing curves for different β's. Smaller than value of β, faster it transitions from supervision loss to consistency loss. In this work, β=0.65 is chosen based on [40].

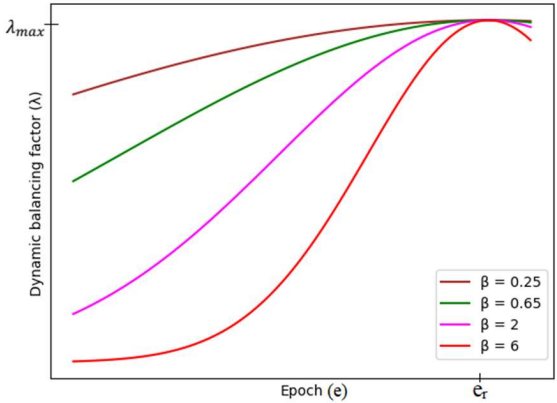

**Fig. 2 Dynamic balancing curves for different values of β**

*4) Inference using Knowledge Distillation*

During testing, given an unlabeled facial image, average of predictions or majority voting from the three networks can be used to infer expression label. However, this would favor the argument that superior results are due to the ensembling effect of the three networks and not due to the effect of CCT framework as such. Further, this would also involve more parameters, and consequently more floating point operations during inference. In fact, it was observed in all our experiments that all the three networks perform equally well, and so arbitrarily any one of the networks can be used during inference. However, to formally have a model with 1/3rd of the training parameters during inference, a simple knowledge distillation [44] scheme is proposed as follows.

Let the trained three networks model be the teacher network T. Let S be the student network that has the same architecture like any of the three networks in T. Obviously, S has 1/3rd of the parameters of T. Let $p_{j,T}$ be the softened output of softmax of the $j^{th}$ network in T with its $i^{th}$ component as $p_{j,T}^i$. The corresponding logits of the $j^{th}$ network are $u_j$. Let $p_S^H$ and $p_S$ be the hard and softened output of softmax of the student network S, respectively. Their corresponding ith components are $p_S^{i,H}$ and $p_S^i$. Let v be the logits of the student network S. Then

$$p_{j,T}^i = \frac{e^{u_j^i/U}}{\sum_{k=1}^N e^{u_j^k/U}}, \quad p_S^i = \frac{e^{v^i/U}}{\sum_{k=1}^N e^{v^k/U}}, \quad p_S^{i,H} = \frac{e^{v^i}}{\sum_{k=1}^N e^{v^k}} \quad (5)$$

where U is the temperature constant used to soften the softmax.

We distill the knowledge from T to S by minimizing the following loss:

$$L_{kd} = 0.5 * \sum_{j=1}^3 L_{CE}(p_{j,T}, p_S) + 0.5 * L_{CE}(p_S^H, y) \quad (6)$$

where $L_{CE}$ is the cross-entropy loss and y is the groundtruth vector. The first term in the above equation enforces the student network to mimic the classification behaviour of all the three networks in the teacher model T. The second term forces the predictions of the student network to be equal to the groundtruth values.

*C. Training Algorithm*
**Algorithm 1:** CCT training algorithm
**Input:** Network **f** with parameters $\{\theta_1, \theta_2, \theta_3\}$, dataset (D), number of epochs ($e_{max}$), maximum λ value ($\lambda_{max}$), learning rate ($\eta$).
**Steps:**
1. Initialize $\theta_1$, $\theta_2$ and $\theta_3$ randomly.
2. **for** e = 1, 2,…, $e_{max}$ epochs do
   i. Sample a minibatch $D_n$ from D
   ii. Compute $p_i = f(x, \theta_i) \; \forall x \epsilon D_n$ ,($1 \leq i \leq 3$)
   iii. Compute dynamic weighting factor λ using Eq. (4)
   iv. Compute joint loss L using Eqs. (1), (2) and (3)
   v. Update $\theta = \theta - \eta \nabla L$
**Output:** Return $\{\theta_1, \theta_2, \theta_3\}$

## IV. EXPERIMENTS

*A. Datasets*

In this section, first the benchmark datasets used for FER in-the-wild settings are described and then our annotated FEC test subset.

*1) In-the-wild FER Datasets*

**AffectNet** [15] is the largest facial expression dataset with 1M images out of which 0.44M are manually annotated and remaining 0.46M images are automatically annotated for the presence of eight (neutral, happy, angry, sad, fear, surprise, disgust, contempt) facial expressions. Since some SOTA methods present results only for the first seven expressions (without contempt) while few others quote for all eight expressions, to compare against the other works, the former case is denoted by AffectNet-7 while the latter by AffectNet-8. AffectNet-7 consists of 3500 images for validation whereas AffectNet-8 has 4000 images. The automatically annotated subset is used for training under real noisy conditions and manually annotated subset for training with synthetic noise. **RAFDB** [17]-[18] contains 29762 facial images tagged with basic or compound expressions by 40 annotators. In this work, the subset with 7-basic emotions consisting of 12,271 images for is used for training and 3068 images for testing. **FERPlus** [16], an extended version of FER2013, consists of 28709 images for training, 3589 images for validation and 3589 for testing with all 8-basic emotions.

In this work, manually annotated test subsets of AffectNet, RAFDB and FERPlus are reffered as clean test subsets of AffectNet, RAFDB and FERPlus, respectively.

### 2) Annotated FEC Test Subset for Expression Recognition

Vemulapalli et al. [27] released a large scale facial expression comparison (FEC) dataset in-the-wild consisting of 500K triplets generated using 157K face images, along with annotations that specify which two expressions in each triplet are most similar than the others. FEC, while rich in its size and diversity, does not provide annotations for basic emotions for FER. A noisy training subset and a clean test subset are created from FEC. For noisy training set, around 86K training images are considered and automatically annotated using SOTA algorithm SCN [19] for 8 basic emotions. Since the number of contempt expressions obtained are only 17, these are not considered. The rest of the images are referred as FEC training set. With regard to clean test subset, 5257 images are annotated as follows. Since expression is naturally perceived by public, crowdsourcing is used as a first step to get labels. Sixteen different annotators were involved. They labeled the images for one of the 8 basic emotions, and none of above if they could not identify the emotion. Motivated by [45]-[46], to arrive at final clean annotation, PM method [45]-[46] is used to infer true expression labels, which takes into account each annotator's expertise. PM is an iterative method which jointly infers annotator's expertise as well as the expression label. Let $v_i^a$ represent the label of $i^{th}$ image by annotator 'a' with expertise $e^a$. Let $v_i^*$ represent its true label. PM method models each annotator's expertise as a single value $e^a \in [0, \infty)$ where higher value means higher expertise. Initially, each annotator 'a' is assigned same expertise $e^a = 1$. Then the following two steps are iterated until convergence:

i) $\quad v_i^* = \arg\max_v \sum_a e^a \cdot \mathbb{1}_{v = v_i^a}$

ii) $\quad e^a = -\log\left(\frac{\Sigma_i \mathbb{1}_{v_i^* \neq v_i^a}}{\max_a \Sigma_i \mathbb{1}_{v_i^* \neq v_i^a}}\right)$

Here, $\mathbb{1}_{\{.\}}$ is the indicator function.

Intuitively, step (i) does majority voting on labels from high expertise annotators (not all annotators) while step (ii) refines the annotators expertise based on step (i). By using this method, true expression labels are inferred for 5257 images out of which 546 are discarded as they belonged to none of the above category. Remaining 4711 images will be referred as FEC test set.

**Meta data of FEC test set:** Precise locations of bounding boxes and expression labels along with URL of these images will be released at https://github.com/1980x/CCT. Also, gender and age groups (5 ranges) for these images are manually annoated. It consists of 48% females and 52% males. Age ranges are between 0-70 years.

### B. Implementation details

Face images are detected and aligned using MTCNN [47], and further resized to 224x224. The individual networks in CCT are ResNet-18 [48] pre-trained on MS-Celeb-1M [49] dataset. Implementation is done in Pytorch. Batch size is set to 256. Optimizer used is Adam. Learning rate (lr) is initialized as 0.001 for base networks and 0.01 for the classification layer. Further, lr is decayed exponentially by a factor of 0.95 every epoch. Data augmentation includes random horizontal flipping, random erasing and color jitter. $\lambda_{max}$ is set to 0.9 and β to 0.65. Oversampling is used to tackle class imbalance based on [50]. Evaluation metric considered is overall accuracy. Inference is based on a simple knowledge distillation. Comparison is done with recent SOTA methods. Note that methods like decoupling [38], Coteaching [25], Coteaching+ [28], JoCoR [39] and NCT [40] also employ multiple networks while training to mitigate the influence of noisy labels. Our codes are available at *https://github.com/1980x/CCT*.

### C. Evaluation of CCT on synthetic noisy data

On RAFDB, FERPlus and AffectNet-8 datasets, 10%, 20%, 30% and 40% labels are randomly changed for training images. CCT is compared with ResNet18 trained with CE loss (referred as standard), Coteaching [25], Coteaching+ [28], JoCoR [39], NCT [40] and SCN [19]. Author's public implementations are used for JoCoR and SCN, and our implementation of NCT since NCT code is not publicly available. Results are presented in Table 1. CCT outperforms all the methods by a significant margin. Specifically, it has atleast over 2% gain on average in comparison to the second best performing method NCT. In fact, as noise rate raises to 40%, CCT gains 4.5% on average over NCT.

### D. Evaluation of CCT on real noisy data

Unlike specific synthetic noise, real-noisy FER datasets can include noise of any level and any type. CCT is trained using automatically labelled subset of AffectNet-7 and tested on AffectNet-7 validation set. A similar procedure is followed for FEC train and test sets. The performance of CCT is compared with methods mentioned in the previous section. CCT is also compared with the recent SOTA methods specifically proposed for FER including SCN [19], RAN [50], GACNN [51] and OADN [52]. In this set of methods specifically proposed for FER, only SCN [19] handles noisy labels. Others do not explicitly handle noisy labels. OADN and GACNN methods are implemented by us while author's public implementation is used for SCN and RAN. Results are shown in Table 2. CCT is superior to all the other methods.

**Table 1: Performance comparison (accuracy %) in presence of synthetic noise on FER datasets**

| Noise level | Method | RAFDB | FERPlus | AffectNet-8 |
|---|---|---|---|---|
| 10 | Standard | 81.74 | 85.87 | 59.27 |
| 10 | Coteaching | 80.18 | 86 | 58.93 |
| 10 | Coteaching+ | 81.84 | 83.33 | 53.73 |
| 10 | JoCR | 84.84 | 85.91 | 58.05 |
| 10 | NCT | 87.42 | 87.28 | 59.70 |
| 10 | SCN | 82.18 | 84.28 | 58.58 |
| 10 | **CCT** | **89.47** | **89.29** | **61.47** |
| 20 | Standard | 79.60 | 84.02 | 58.12 |
| 20 | Coteaching | 79.56 | 85.5 | 57 |
| 20 | Coteaching+ | 81.12 | 76.44 | 49.55 |
| 20 | JoCR | 82.79 | 83.71 | 57.28 |
| 20 | NCT | 85.29 | 86.42 | 59.28 |
| 20 | SCN | 80.01 | 83.17 | 57.25 |
| 20 | **CCT** | **87.51** | **88.05** | **60.53** |
| 30 | Standard | 74.31 | 82.30 | 56.50 |
| 30 | Coteaching | 75 | 83.48 | 54.22 |
| 30 | Coteaching+ | 80.05 | 75.83 | 44.9 |
| 30 | JoCR | 80.96 | 81.51 | 54.38 |
| 30 | NCT | 82.66 | 83.93 | 56.23 |
| 30 | SCN | 77.46 | 82.47 | 55.05 |
| 30 | **CCT** | **86.44** | **87.28** | **57.65** |
| 40 | Standard | 70.30 | 80.36 | 54.65 |
| 40 | Coteaching | 61.18 | 80.52 | 50.45 |
| 40 | Coteaching+ | 80.05 | 75.83 | 44.9 |
| 40 | JoCR | 80.96 | 81.51 | 54.38 |
| 40 | NCT | 79.01 | 81.86 | 52.25 |
| 40 | **CCT** | **84.22** | **86.13** | **55.58** |

**Table 2: Performance comparison (accuracy %) on real noisy datasets**

| Method | AffectNet-7 | FEC |
|---|---|---|
| Standard | 53.85 | 53.93 |
| Coteaching | 52.37 | 54.36 |
| Coteaching+ | 55.08 | 56.63 |
| JoCR | 55.00 | 54.59 |
| NCT | 56.46 | 57.16 |
| RAN | 56.43 | 53.51 |
| GACNN | 52.43 | 53.13 |
| OADN | 55.37 | 51.22 |
| SCN | 56.03 | 52.96 |
| **CCT** | **57.00** | **57.90** |

The individual expression discrimination ability of CCT is also analysed on the real noisy data. The confusion matrices with respect to AffectNet-7 and FEC test sets are shown in Fig. 3. In both the cases, happiness is the easiest recognizable expression followed by neutral and surprise. Fear is highly confused with surprise in both the cases. Similarly, anger is significantly confused with neutral. While disgust is the most difficult expression to recognize in AffectNet-7, sadness is relatively difficult to recognize in the FEC test set. Note that the performance in the FEC test set is poor in comparison to performance on AffectNet-7 because FEC test set is significantly larger than AffectNet-7 test set (see section IV.A.2), and that the training set of AffectNet-7 is much larger than FEC train set. This aspect (larger test size and relatively smaller training size than AffectNet which is the largest FER in-the-wild dataset) of our curated FEC set, we believe, makes our curated FEC set a challenging set and useful for the research community.

**Table 3: Performance comparison (accuracy %) against methods that deal with noisy labels on clean test subsets**

| Method | RAFDB | FERPlus | AffectNet-7 | AffectNet-8 |
|---|---|---|---|---|
| Standard | 85.07 | 87.63 | 63.54 | 59.72 |
| JoCoR | 86.08 | 86.8 | 61.05 | 60.68 |
| NCT | 88.2 | 88.3 | 64.91 | 62 |
| SCN | 88.14 | 89.35 | 64.2 | 60.23 |
| **CCT** | **90.84** | **89.99** | **66** | **62.65** |

*E. Evaluation of CCT on clean test subsets*

The performance of CCT is compared on the clean test subsets of RAFDB, FERPlus and AffectNet datasets (the noisy version has been covered in Tables 1 and 2). First, it is compared with methods that deal with noisy labels in Table 3. CCT shows around at least 2% gain over all the methods in all the databases. Next, CCT is also compared with other SOTA methods (that do not handle noisy labels) on the clean test subsets of RAFDB (see Table 4), FERPlus (see Table 5) and AffectNet (see Tables 6 and 7) datasets. CCT reports the newest SOTA performance on all the databases. Particularly, CCT stands out as the first method to breach the 90% mark on the RAFDB dataset, reporting 90.74% accuracy. This is almost 1% more than the next best performing OADN and at least 5.5% more than a recent method called GACNN. Similarly, in FERPlus, CCT has almost breached 90%. It has almost 5% advantage over GACNN and a slightly better performance than the current SOTA method on FERPlus, namely GCN. In AffectNet datasets, CCT again displays at least 1.5% to 2% performance gain over the next best performing method GCN. These newest SOTA results of CCT demonstrate the CCT's utility as a general purpose robust FER learning framework.

The confusion matrices of CCT approach with regard to clean test subsets of RAFDB, AffectNet-7 and FERPlus are shown in Fig. 4. As with real noisy case, happiness is the easiest recognizable expression across all the clean test subsets, followed by neutral and surprise expressions. Fear is generally confused with surprise. Further, disgust is confused with anger in FERPlus and AffectNet-7 clean test subsets, and with sad expression in RAFDB. Disgust is a difficult expression to recognize across all clean subsets. However, in FERPlus, contempt is the most difficult expression, exhibiting high ambiguity with neutral. In general, CCT performs consistently in discriminating expressions in both real noisy and clean cases, and across all the datasets. This substantiates its consistent top performance on all the datasets.

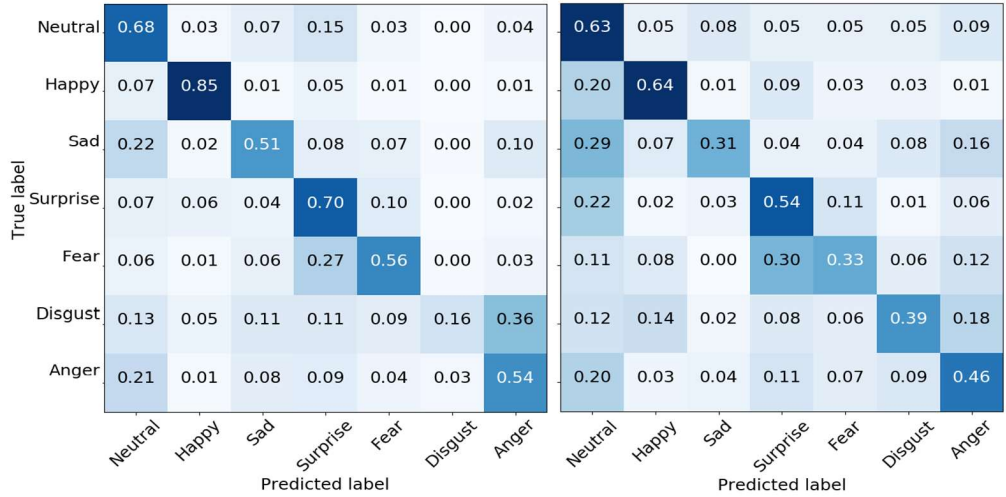

**Fig. 3** Confusion plots for AffectNet-7 (left) and our annotated FEC test subset (right)

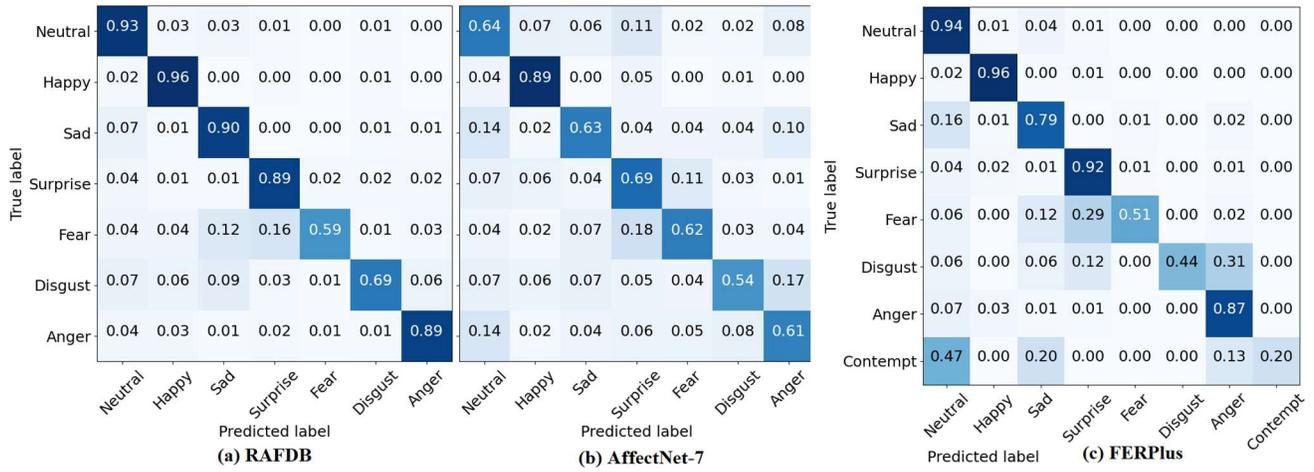

**Fig. 4** Confusion plots for clean test subsets

**Table 4:** Comparison against recent SOTA methods that do not deal with noisy labels on clean test subset of RAFDB

| Method | Year | Accuracy (%) |
|---|---|---|
| GACNN [51] | 2018 | 85.07 |
| Covariance pooling [54] | 2019 | 87.00 |
| IPA2LT [41] | 2020 | 86.77 |
| RAN [50] | 2020 | 86.9 |
| THIN [55] | 2020 | 87.81 |
| Mahmoudi et al. [56] | 2020 | 88.02 |
| Liu et al. [57] | 2020 | 88.02 |
| GCN [58] | 2020 | 89.41 |
| OADN [52] | 2020 | 89.83 |
| **CCT** | - | **90.84** |

**Table 5:** Comparison against recent SOTA methods that do not deal with noisy labels on clean test subset of FERPlus (* denotes our implementation)

| Method | Year | Accuracy (%) |
|---|---|---|
| GACNN [51] | 2018 | 84.86* |
| Georgescu et al. [59] | 2019 | 87.76 |
| RAN [50] | 2020 | 89.26 |
| OADN [52] | 2020 | 88.71* |
| GCN [58] | 2020 | 89.39 |
| ESR-9 [60] | 2020 | 87.15 |
| **CCT** | - | **89.99** |

**Table 6: Comparison against recent SOTA methods that do not deal with noisy labels on clean test subset of AffectNet-7 (* denotes our implementation)**

| Method | Year | Accuracy (%) |
|---|---|---|
| Annotators Agreement [15] | 2017 | 65.3 |
| Kollias et al.[61] | 2018 | 60.0 |
| GACNN[51] | 2018 | 58.78 |
| HERO[62] | 2019 | 62.11 |
| Yongjian et al. [63] | 2019 | 62.7 |
| FMPN[64] | 2019 | 61.5 |
| RAN[50] | 2020 | 61.71* |
| Georgescu et al. [59] | 2019 | 63.31 |
| IPA2LT [41] | 2020 | 57.31 |
| OADN[52] | 2020 | 64.06 |
| THIN[55] | 2020 | 63.97 |
| GCN[58] | 2020 | 64.46 |
| **CCT** | - | **66.0** |

**Table 7: Comparison against recent SOTA methods that do not deal with noisy labels on clean test subset of AffectNet-8 (* denotes our implementation)**

| Method | Year | Accuracy (%) |
|---|---|---|
| GACNN[51] | 2018 | 55.05* |
| Georgescu et al. [59] | 2019 | 59.58 |
| RAN [50] | 2020 | 59.50 |
| OADN [52] | 2020 | 58.92* |
| GCN [58] | 2020 | 60.58 |
| ESR-9 [60] | 2020 | 59.3 |
| **CCT** | - | **62.65** |

**Table 8: Evaluation with and without consistency loss using 1 - 4 networks in CCT (accuracy %)**

| Dataset | Noise | Loss | | Number of networks | | | |
|---|---|---|---|---|---|---|---|
| | | $L_{sup}$ | $L_{con}$ | 1 | 2 | 3 | 4 |
| RAFDB | 0% | ✗ | ✓ | 85.07 | 89.96 | 89.92 | 89.86 |
| | | ✓ | ✓ | NA | 90.09 | **90.84** | 89.86 |
| | 40% | ✗ | ✓ | 70.3 | 80.7 | 80.31 | 82 |
| | | ✓ | ✓ | NA | 82.72 | 84.22 | **84.84** |
| FERPlus | 0% | ✗ | ✓ | 87.63 | 89.16 | 89.45 | 89.58 |
| | | ✓ | ✓ | NA | 89.7 | **89.99** | 87.25 |
| | 40% | ✗ | ✓ | 80.36 | 82.69 | 83.04 | 83.64 |
| | | ✓ | ✓ | NA | 85.08 | 86.13 | **86.16** |

### F. Further Analysis

#### 1) Impact of consistency loss

The effect of consistency loss in CCT is evaluated in Table 8. In the presence of noise, consistency loss enhances performance by around 3 to 4%, clearly indicating that noisy labels have been handled much better with consistency loss than not having it. Without consistency loss, each network becomes independent with only associated supervision loss to guide them. This makes each network memorize noisy labels over training. During knowledge distillation, the memory is passed over to the student, thereby negatively impacting the generalizability of the network. Hence, consistency loss plays the crucial role to mitigate the influence of noise and enhance the performance.

#### 2) Number of networks

Table 8 also shows the influence of number of networks that are collaboratively trained in CCT. It can be observed that model with 4 networks performs the best in the presence of noise. This is because of the region where the 4 networks come into consensus is relatively smaller, thereby avoiding noisier labels during training. However, with 4 networks, the number of parameters also significantly increases. Note that, in the presence of noise, the 3 networks model has almost similar performance to 4 networks model, and is definitely much better than the 2 networks model. Hence, keeping number of parameters in mind, the 3 networks model is preferred.

#### 3) Effect of temperature constant U

The effect of temperature constant U in knowledge distillation for inference is also analysed. The plot is available in Fig. 5. The best performance is obtained for U = 2. Making the target softer with high U brings down the performance. Softer target, as proposed in [44], is likely to carry a lot of helpful information which cannot be encoded in a single hard target. However, since the three networks in the teacher model T are collaboratively trained with consistency loss, they contain relevant information, uninfluenced by the noisy annotations. Hence, minimal softening would suffice to pass on the information to the student network S. Higher values of U are likely to distort the information.

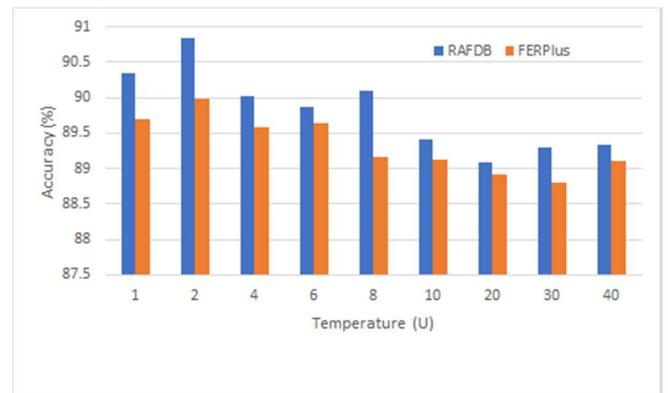

**Fig. 5 Influence of temperature constant in Knowledge distillation**

#### 4) Summary of techniques used in CCT

Table 9 compares the techniques used in CCT against other SOTA methods. CCT does not rely only on low loss samples.

Table 9 Comparison of proposed method with related approaches

| Method | Decoupling | Coteaching | Coteaching+ | JoCoR | NCT | CCT |
|---|---|---|---|---|---|---|
| Small loss | ✗ | ✓ | ✓ | ✓ | ✗ | ✗ |
| Cross update | ✗ | ✓ | ✓ | ✗ | ✗ | ✗ |
| Disagreement | ✓ | ✗ | ✓ | ✗ | ✗ | ✗ |
| Agreement | ✗ | ✗ | ✗ | ✓ | ✗ | ✓ |
| Noise rate dependence | ✓ | ✓ | ✓ | ✓ | ✗ | ✗ |
| Joint update | ✗ | ✗ | ✗ | ✓ | ✗ | ✓ |
| Dynamic weighing | ✗ | ✗ | ✗ | ✗ | ✓ | ✓ |

This helps CCT not to miss out on hard instances, which is substantiated by the performance it has reported. Unlike [28] and [38], CCT banks on agreement principle wherein it forces consensus among predictions of all the three networks through the consistency loss. Supervision loss helps in the early part of the training to learn clean instances but CCT dynamically shifts to consistency loss in the later part of the training to suppress the influence of noisy instances. CCT does not require cross-updates as co-training provides the advantages of cross-updates.

*5) Visualization*

The salient regions focused upon by the distilled network for inference are also visualized. Grad-CAM [53] is used to obtain activation maps on the final layer of the trained student ResNet-18 model. The visualizations are shown in Fig. 6. Red color denotes higher saliency while blue color denotes lower saliency. It can be noted that, though there are occlusions like glasses, hat, hand etc., the model avoids them and focuses on non-occluded parts to infer the emotion. The model is also able to handle pose variations as depicted in the figure. Further, for the neutral expression, moderately salient pixels suffice to infer it. Anger can be subtle as in the first image in second row, or expressive as in the second image in the first row. In the former case, the model relies on the eye region to infer the emotion. In the latter case, mouth region provides the cue.

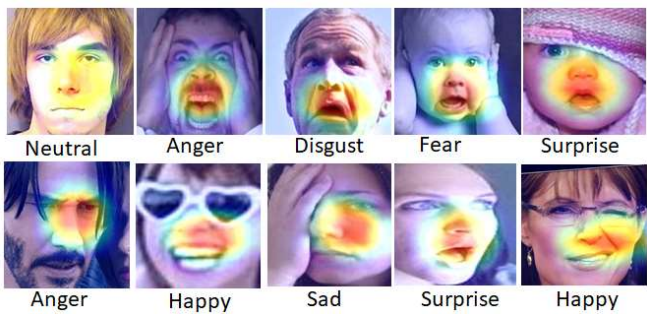

**Fig. 6 Activation maps using GRAD-CAM on sample images from RAFDB and AffectNet datasets (predicted expressions are displayed below each image). Red color denotes higher saliency while blue color denotes lower saliency. Yellow color denotes moderate saliency.**

Other expressions bank on the central part of the face. Fear also has some dependence on the eye region.

## V. CONCLUSIONS

In this work, a robust framework called CCT is proposed for effectively training a FER system with noisy annotations. CCT combated the noisy labels by co-training three networks. Supervision loss at early stage of training and gradual transition to consistency loss at latter part of training ensured that the noisy labels did not influence the training. Both the losses are balanced dynamically. Inference is done through a single network distilled from co-trained three networks. This work also presents an annotated test subset of data that will be released publicly which could further research in FER under noisy labels. Extensive experiments on three widely used FER datasets and the annotated test subset demonstrated the effectiveness of CCT. Infact, CCT reports newest SOTA results on the standard FER datasets. An extension to this work could focus on curing the noisy labels during training, wherever possible.

## ACKNOWLEDGMENT

This work is dedicated to Founder Chancellor of Sri Sathya Sai Institute of Higher learning, Bhagavan Sri Sathya Sai Baba. Authors are also thankful to the annotators who annotated the dataset.